\documentclass[final]{article}

\usepackage{neurips_2025}

\usepackage[utf8]{inputenc} % allow utf-8 input
\usepackage[T1]{fontenc}    % use 8-bit T1 fonts
\usepackage{hyperref}       % hyperlinks
\usepackage{url}            % simple URL typesetting
\usepackage{booktabs}       % professional-quality tables
\usepackage{amsfonts}       % blackboard math symbols
\usepackage{nicefrac}       % compact symbols for 1/2, etc.
\usepackage{microtype}      % microtypography
\usepackage{xcolor}         % colors
\usepackage{booktabs}
\usepackage{graphicx}
\usepackage{amsmath}
\usepackage{float}
\usepackage{array}
\newtheorem{theorem}{Theorem}[section]

\newtheorem{lemma}[theorem]{Lemma}
\newtheorem{definition}{Definition}[section]
\newtheorem{assumption}{Assumption}[section]
\usepackage{caption}
\title{HOME-3: High-Order Momentum Estimator with Third-Power Gradient for Convex and Smooth Nonconvex Optimization}

\author{%
  Wei Zhang\thanks{The corresponding author.}\\
  School of Computer and Cyber Sciences \\
  Augusta University, Augusta, GA, USA  \\
  \texttt{wzhang2@augusta.edu} \\
  \And
  Arif Hassan Zidan \\
  School of Computer and Cyber Sciences \\
  Augusta University, Augusta, GA, USA \\
  \texttt{azidan@augusta.edu} \\
  \And
   Afrar Jahin \\
   School of Computer and Cyber Sciences\\
   Augusta University, Augusta, GA, USA\\
  \texttt{ajahin@augusta.edu} \\
  \And
  Yu Bao \\
  Department of Graduate Psychology \\
  James Madison University, Harrisonburg, VA, USA \\
  \texttt{bao2yx@jmu.edu} \\
  \And
  Tianming Liu \\
  School of Computing \\
  University of Georgia, Athens, GA, USA \\
  \texttt{tliu@uga.edu} \\
}

\begin{document}

\maketitle

\begin{abstract}
  Momentum-based gradients are essential for optimizing advanced machine learning models, as they not only accelerate convergence but also advance optimizers to escape stationary points. While most state-of-the-art momentum techniques utilize lower-order gradients, such as the squared first-order gradient, there has been limited exploration of higher-order gradients, particularly those raised to powers greater than two. In this work, we introduce the concept of high-order momentum, where momentum is constructed using higher-power gradients, with a focus on the third-power of the first-order gradient as a representative case. Our research offers both theoretical and empirical support for this approach. Theoretically, we demonstrate that incorporating third-power gradients can improve the convergence bounds of gradient-based optimizers for both convex and smooth nonconvex problems. Empirically, we validate these findings through extensive experiments across convex, smooth nonconvex, and nonsmooth nonconvex optimization tasks. Across all cases, high-order momentum consistently outperforms conventional low-order momentum methods, showcasing superior performance in various optimization problems.
\end{abstract}

\section{Introduction}
Optimization problems in machine learning are commonly tackled using gradient-based optimizers, which rely on either full gradients—computed from the entire dataset—or stochastic gradients, derived from mini-batches. While full gradients guarantee eventual convergence, stochastic gradients offer enhanced computational efficiency~\citep{Hazan2007logarithmic, Nemirovski2009robuststochastic, Rakhlin2011stronglyconvex}. Over the past decade, research has shown that combining full gradients, stochastic gradients, noisy stimuli, batch strategies, sampling, and momentum techniques in gradient-based optimizers can lead to favorable convergence, expected accuracy, and improved robustness~\citep{shalev2013stochasticdual, Zhang2012cimmuncation, Johnson2013accelerating, Defazio2014saga, Arjevani2015complexityl, Lin2015universal, Allen2017firstdirect, haji2021comparison}.

Momentum, one of the most influential techniques, is widely used in gradient-based optimizers to further improve performance~\citep{liu2020improved, loizou2020momentum, haji2021comparison}. Intuitively, momentum addresses the issue of slow convergence in later stages of optimization, such as near $(\delta,\epsilon)$-Goldstein stationary points \citep{clarke1974necessary, clarke1975generalized, clarke1981generalized, clarke1990optimization, jordan2023deterministic}, where gradients oscillate within a narrow range. Momentum helps by driving gradients away from these oscillations and toward the global optimum, making it especially effective for nonsmooth nonconvex objectives, such as those found in Deep Neural Networks (DNNs)~\citep{mai2020convergence, wang2021distributed, wang2022proximal, jordan2023deterministic}. 

Due to these advantages, leading optimizers like Adam, STORM, and $STORM^+$~\citep{Kingma2014adam, cutkosky2019momentum, levy2021storm+} incorporate momentum to achieve higher accuracy and reduce the likelihood of getting trapped in stationary points. For instance, Adam uses two momentum terms—first-order and squared first-order gradients—to optimize objective functions, often outperforming alternatives like AdaGrad and SGD~\citep{Kingma2014adam, lydia2019adagrad, chandra2022gradient, beznosikov2023stochastic}. STORM, which uses a stochastic recursive momentum term based on squared gradients, has been shown to achieve better accuracy than Adam when optimizing ResNet~\citep{cutkosky2019momentum}, and the more recent $STORM^+$ enhances this approach with adaptive learning rates, eliminating the need for parameter tuning~\citep{levy2021storm+}.

While first-order and squared gradients dominate current momentum-based approaches, exploring higher-order momentum holds great potential. For instance, incorporating third-power gradients could further enhance the convergence bound of gradient-based optimizers. In this work, we introduce the High-Order Momentum Estimator (HOME) optimizer, a framework designed to explore and advance high-order momentum techniques. Our focus is on \textit{HOME}-3, which leverages third-power gradients to enhance momentum, such as $(f^{\prime})^3$. First, we present a theoretical analysis showing that \textit{HOME}-3 significantly improves convergence bounds for both convex and smooth nonconvex optimization problems. We then extend our numerical experiments to nonsmooth nonconvex problems, where \textit{HOME}-3 consistently outperforms other momentum-based optimizers. Finally, we use statistical techniques to quantify the performance of \textit{HOME}-3, validating both the effectiveness and robustness of third-power gradients in momentum.

\noindent \textbf{Contributions}: In this work, the potential contributions of \textit{HOME} are categorized as follows:

\textit{Third-Order Momentum Enhances Convergence Bound in Convex Optimization} (\textbf{Theorem}~\ref{theorem1}): Based on the assumptions and properties of convex objective functions (see \textbf{Assumption} \ref{assumption1}), the proposed \textit{HOME}-3 optimizer enhances the convergence bound to $O(1 \slash T^{5 \slash 6})$. Detailed proof of \textbf{Theorem} \ref{theorem1} can be viewed in Appendix A of the Supplementary Material.

\textit{Third-Order Momentum Advances Convergence Bound in Smooth Nonconvex Problems} (\textbf{Theorem}~\ref{theorem2}): According to the assumptions and properties of smooth nonconvex functions (see \textbf{Assumption} \ref{assumption2}), the \textit{HOME}-3 optimizer advances the convergence bound to approximately $O(1 \slash T^{5 \slash 6})$. The proof for \textbf{Theorem} \ref{theorem2} is provided in Appendix A of the Supplementary Material.

\textit{Third-Order Momentum Enhances Convergence for Nonsmooth Nonconvex Problems} (\textbf{Theorem}~\ref{theorem3}): We empirically investigate the performance of high-order momentum optimizers on nonsmooth nonconvex problems, as illustrated in Figure~\ref{fig:fig3}. To further validate the performance of \textit{HOME}-3, we employ a deep neural network, since the objective function of a multi-layer deep neural network is typically nonsmooth and nonconvex~\citep{jordan2023deterministic}. The results, shown in Figures~\ref{fig:fig3} and~\ref{fig:fig6}, indicate that \textit{HOME}-3 outperforms other peer momentum-based optimizers. Additionally, we explore the advantages of coordinate randomization in \textbf{Lemma}~\ref{lemma1} and\textbf{Theorem}~\ref{theorem3}, demonstrating that it preserves the convergence bound of the original gradient-based optimizer.

\noindent \textbf{Related Work}: In the field of convex and smooth nonconvex optimization, Kingma's work on Adam~\citep{Kingma2014adam} demonstrated that momentum, built on the first-order and squared gradients, can achieve a convergence bound of $O(1 \slash T^{1 \slash 2})$ for convex problems. Similarly, STORM, which uses a recursive stochastic momentum, obtains a convergence bound of $O(1 \slash T^{1 \slash 3})$ for smooth nonconvex problems~\citep{cutkosky2019momentum}. More recently, $STORM^+$ achieved a convergence bound of $O(1 \slash T^{1 \slash 2} + \sigma^{1 \slash 3} \slash T^{1 \slash 3})$ \citep{levy2021storm+}.

\section{Preliminaries: Definitions and Assumptions}
\label{gen_inst}

% \begin{table}[h]
% \centering
% \caption{The definitions of mathematical symbols.}
% \label{tab:symbols}
% \begin{tabular}{>{\centering\arraybackslash}m{4cm} >{\centering\arraybackslash}m{10cm}}
% \toprule
% \textbf{Symbol} & \textbf{Description} \\
% \midrule
% \( f(x) \) & Objective function to minimize \\
% \( x_t \in \mathbb{R}^D \) & Optimization variable at iteration \( t \) \\
% \( \nabla f(x) \) & Gradient of \( f \) at point \( x \) \\
% \( g_t = \nabla f(x_t) \) & Gradient at iteration \( t \) \\
% \( g_t^n \) & Elementwise \( n \)-th power of gradient \\
% \( M_t \) & Momentum term at iteration \( t \) \\
% \( V_t \) & Second-moment term (squared gradients) \\
% \( S_t \) & Third-moment term (cubed gradients) \\
% \( \alpha \) & Learning rate \\
% \( \mathcal{G} \) & Gradient-based update operator \\
% \( \mathcal{R} \) & Coordinate randomization operator \\
% \( \hat{x}_t \) & Randomized output after applying \( \mathcal{R} \) \\
% \( D \) & Dimension of input space \\
% \( T \) & Total number of iterations \\
% \( \epsilon \) & Small threshold for stationarity \\
% \bottomrule
% \end{tabular}
% \end{table}

We begin by formalizing the optimization problem and providing key definitions and assumptions that form the theoretical foundation of this work. We focus on analyzing first-order gradient-based methods that incorporate higher-order momentum. All important mathematical symbols can be viewed in Table~\ref{table:table3} in Appendix A, Supplementary Material.

\subsection{Problem Setup}

Let \( f: \mathbb{R}^D \rightarrow \mathbb{R} \) be a real-valued objective function defined over a \( D \)-dimensional Euclidean space, where \( D < \infty \). We consider the following unconstrained minimization problem:
\begin{equation}
\label{eq:opt}
    \min_{x \in \mathbb{R}^D} f(x)
\end{equation}
Depending on the properties of \( f \), the problem may be convex, smooth nonconvex, or nonsmooth nonconvex. In this work, our theoretical analyses are primarily concerned with convex and smooth nonconvex settings. For nonsmooth nonconvex problems, we conduct empirical investigations and leverage recent advances in coordinate randomization~\citep{zhang2022sadam}.

\subsection{Definitions}

\begin{definition}
\label{def1}
([High-Order Momentum) Let \( f: \mathbb{R}^D \to \mathbb{R} \) be a differentiable function, and let \( \nabla f(x) = \left[ \partial_1 f(x), \dots, \partial_D f(x) \right]^T \in \mathbb{R}^D \) be its gradient. The \emph{high-order momentum} vector \( M \in \mathbb{R}^D \) of order \( n \in \mathbb{N} \) at point \( x \) is defined component-wise as:
\[
M_i = \sum_{k=1}^n \beta_k \left( \partial_i f(x) \right)^k, \quad \text{for } i = 1, \dots, D
\]
where \( \beta_k \in \mathbb{R} \) are scalar hyperparameters.
\end{definition}

\begin{definition}
\label{def2}
(Smoothness) A differentiable function \( f: \mathbb{R}^D \to \mathbb{R} \) is said to be \( k \)-times continuously differentiable and \( L \)-smooth of order \( k \) if for all \( x, y \in \mathbb{R}^D \), the \( k \)-th derivative satisfies:
\[
\| \nabla^k f(x) - \nabla^k f(y) \| \leq L \| x - y \|,
\]
where \( \nabla^k f(x) \) denotes the \( k \)-th order derivative tensor and \( \| \cdot \| \) is the Euclidean norm.
\end{definition}

% \begin{definition}
% \label{def3}
% (Gradient-based Optimization Operator) Given an operator as $\mathcal{G}: \mathbb{R} \rightarrow \mathbb{R}^D$, $\mathcal{G}$ denotes a gradient-based optimization operator. For example, suppose $t(\forall t \in \mathbb{N})$ as current iteration, we have $\mathcal{G} \cdot f(x_t) = x_t - \alpha \cdot \nabla f(x_t)$, the operator $\mathcal{G}$ denotes a first-order gradient-based optimizer.
% \end{definition}

\begin{definition}
\label{def3}
(Gradient-based Operator) Let \( \mathcal{G} \) be a gradient-based update operator acting on a differentiable function \( f: \mathbb{R}^D \to \mathbb{R} \). For a given iterate \( x_t \in \mathbb{R}^D \), the update is defined as:
\[
x_{t+1} = \mathcal{G}(x_t) := x_t - \alpha \cdot \nabla f(x_t),
\]
where \( \alpha > 0 \) is the step size.
\end{definition}

\begin{definition}
\label{def4}
(Coordinate Randomization) Given an operator $\mathcal{R}$ denoted as $\mathcal{R}: \mathbb{R}^D \rightarrow \mathbb{R}^D$, we have $\mathcal{R} \cdot \lbrace x_1, x_2, \cdots, x_D \rbrace = \lbrace \hat{x}_1, \hat{x}_2, \cdots, \hat{x}_D \rbrace$. The operator $\mathcal{R}$ is a coordinate randomization. 
\end{definition}

\begin{definition}
\label{def5}
(Iterative Format of Gradient and Permutation Randomization Operators) Given gradient and permutation randomization operators $\mathcal{G}$ and $\mathcal{R}$, suppose the current iteration as $t$, $\mathcal{G}^t f(x)$ and $\mathcal{R}^t x$ represent an iterative format of gradient and permutation randomization operator within $t$ iterations. For example, $\mathcal{G}^2 f(x) = \mathcal{G} \cdot \mathcal{G} \cdot f(x)$ and $\mathcal{R}^2 x = \mathcal{R} \cdot \mathcal{R} \cdot x$.
\end{definition}

\begin{definition}
\label{def6}
(Initialization and Stationary Point) We denote $x_0$ as an initialized variable for a gradient-based optimizer to begin iteration. Meanwhile, a stationary point is represented by $x_T$, and $T$ indicates the maximum iteration.
\end{definition}

\begin{definition}
\label{def7}
(Iterative Output of Gradient and Coordinate Randomization Operators) Given gradient and permutation randomization operators $\mathcal{G}$ and $\mathcal{R}$, suppose the current iteration as $t$, $\mathcal{G}^t f(x)$ and $\mathcal{R}^t x$ represent gradient and permutation randomization operator within $t$ iterations. The iterative output of gradient and permutation randomization operators are denoted as $x_t=\mathcal{G} \cdot f(x_{t-1})=\mathcal{G}^t \cdot f(x_0)$ and $\hat{x}_t=\mathcal{R} \cdot \mathcal{G} \cdot f(x_{t-1})=\mathcal{R}^t \cdot \mathcal{G}^t \cdot f(x_0)$.
\end{definition}

\subsection{Assumptions}

\noindent Moreover, three vital assumptions are provided below to benefit theoretical analyses of \textit{HOME}-3 optimizer on convex, smooth nonconvex, and nonsmooth nonconvex optimization.
% \begin{assumption}
%     \label{assumption1} 
%     (Convex Assumption) $f(y) \geq f(x) + (\nabla f(x))^T (y-x)$, $x, y \in \mathbb{R}^D$
% \end{assumption}

\begin{assumption}
\label{assumption1} 
(Convex Assumption) The function \( f: \mathbb{R}^D \to \mathbb{R} \) is convex, i.e., for all \( x, y \in \mathbb{R}^D \),
\[
f(y) \geq f(x) + \nabla f(x)^T (y - x).
\]
\end{assumption}

\begin{assumption}
    \label{assumption2} 
    (Smooth Nonconvex Assumption) $f(y) \leq f(x) + (\nabla f(x))^T (y-x) + \frac{L}{2} \cdot \left \| x-y \right \|$, $x, y \in \mathbb{R}^D, L \in \mathbb{R}, L>0$
\end{assumption}

\begin{assumption}
    \label{assumption4} 
    (Continuity of Linear Gradient Composition) Considering iteration from $1$ to $T$, for any $t \in \lbrack 1,T \rbrack$, and $n \in \mathbb{N}$ as the power for gradient, $\forall \epsilon>0$, the following equation holds:
    \begin{equation} \label{eq2}
    \begin{gathered}
    \left \| g^n - (k_1g_1^n+k_2g_2^n+\cdots+k_Tg_T^n) \right \| < \epsilon
    \end{gathered}
    \end{equation}
    $\lbrace k_1, k_2, \cdots, k_T \rbrace$ are constant and $\lbrace g_1, g_2, \cdots, g_T \rbrace$ represents first-order gradient in $1, 2, \cdots, T$ iteration.
\end{assumption}

\textbf{Assumption}~\ref{assumption4} facilitates the analyses of convergence bound of \textit{HOME}-3.

\section{Method: High-Order Momentum Estimator (\textit{HOME})}
\label{headings}
This section outlines the details of the \textit{HOME} optimizer, as summarized in Table \ref{table:table1}. At its core, the \textit{HOME} optimizer offers a framework for incorporating high-power first-order gradients to generate high-order momentum. In particular, we focus on analyzing the properties of high-order momentum using a third-power first-order gradient as a starting point and extend our theoretical analysis to even higher-order momenta, such as those utilizing a sixth-power gradient. To facilitate both implementation and validation against other state-of-the-art optimizers, we base our framework on the widely used Adam optimizer. However, in contrast to Adam, which is dominated by first- and second-order momentum terms, our proposed method introduces an innovative update rule that is driven by the interaction between the first and third momentum terms, as shown below:

\begin{equation} \label{eq3}
\begin{gathered}
x_t \leftarrow x_{t-1} - \alpha_t \cdot (\hat{M}_{t-1} - \hat{S}_{t-1}) \slash (\sqrt{\hat{V}_{t-1}}+\epsilon_1)
\end{gathered}
\end{equation}

In~\eqref{eq3}, $\hat{M_t}$, $\hat{V}_t$, and $\hat{S}_t$ denote the first-order, second-order, and third-order momentum (please refer to \textbf{Definition} \ref{def1}). Meanwhile, $\alpha_t$ denotes an adaptive learning rate~\citep{huang2021super}. And $\epsilon_1$ is set the same as Adam~\citep{Kingma2014adam}. In addition, the third momentum term $\hat{S}_t$ is cultivated on the third-power first-order gradient:
\begin{equation} \label{eq4}
\begin{gathered}
S_t \leftarrow \beta_3 S_{t-1} + (1-\beta_3)g_t^3\\
\hat{S}_t \leftarrow \frac{S_t}{1-\beta_3^t}
\end{gathered}
\end{equation}
where $\beta_3$ is an exponential decay and $g_t^3$ represents a third-power gradient within iteration $t$. Intuitively, a higher-power gradient dominates the update when the gradient norm is sufficiently large at the early stage. Otherwise, a lower-order gradient is in charge of the update when the gradient norm is reduced to a small value. That is, the convergence bound of the \textit{HOME} optimizer is adaptive. In addition, other efficient techniques are included for the \textit{HOME} optimizer, such as adaptive learning rate~\citep{huang2021super} and coordinate randomization~\citep{zhang2022sadam} since these techniques guarantee an influential impact~\citep{huang2021super, jordan2023deterministic} on complex optimization, e.g., nonsmooth$\slash$smooth nonconvex problems.

The input for \textit{HOME}-3 optimizer is: $t$ represents current iteration; $T$ defines the maximum iteration; $\alpha_t$ denotes an adaptive step size based on current iteration~\citep{huang2021super}, such as $0.001 \times (1-\frac{t}{T})$; $\beta_1=0.9$, $\beta_2=0.999$, $\beta_3=0.99$ are exponential decay for three momentum terms~\citep{Kingma2014adam}, respectively; currently, $\beta_3$ is manually set, ensuring that $\beta_1 < \beta_3 < \beta_2$; $M_0$ denotes the first-moment vector and initializes as 0; $V_0$ denotes the second momentum vector and is initialized as 0; $S_0$ denotes the third momentum vector and is initialized as 0;  $\epsilon_1$ defines the same in Adam; $\epsilon_2$ represents a threshold when gradient within a stationary point. In this work, we set $\epsilon_2$ the same as $\epsilon_1$.

% \begin{table}[t]
%   \caption{The Pseudo Code of High-Order Momentum Estimator (HOME)}
%   \centering
%   \begin{tabular}{l}
%     \toprule
%    \textbf{Algorithm 1:} \textit{HOME}-3\\
%     \midrule
%     1: \textbf{while} $t<T$\\
%     2:\,\, $g_t \leftarrow \nabla_x f(x_t)$\\
%     3:\,\, $M_t \leftarrow \beta_1 M_{t-1} + (1-\beta_1)g_t$\\
%     4:\,\, $V_t \leftarrow \beta_2 V_{t-1} + (1-\beta_2)g_t^2$\\
%     5:\,\, $S_t \leftarrow \beta_3 S_{t-1} + (1-\beta_3)g_t^3$\\
%     6:\,\, $\hat{M}_t \leftarrow \frac{M_t}{1-\beta_1^t}$\\
%     7:\,\, $\hat{V}_t \leftarrow \frac{V_t}{1-\beta_2^t}$\\
%     8:\,\, $\hat{S}_t \leftarrow \frac{S_t}{1-\beta_3^t}$\\
%     9:\,\, $x_{t+1} \leftarrow x_t - \alpha_t \cdot (\hat{M}_t - \hat{S}_t) \slash (\sqrt{\hat{V}_t}+\epsilon_1)$\\
%     10:\,\, \textbf{if} $\left \| \hat{M}_t-\hat{S}_t\right \| < \epsilon_2$\\
%     11:\,\,\,\, $\hat{x}_{t+1} \leftarrow \mathcal{R}(x_{t+1})$\\
%     12:\,\,\,\, $x_{t+1} \leftarrow \hat{x}_{t+1}$\\
%     13:\,\, \textbf{End if}\\
%     14:\,\, $\textit{t} \leftarrow \textit{t}+1$\\
%     15: \textbf{End while}\\
%     \bottomrule
%   \end{tabular}
%   \label{table:table1}
% \end{table}
 
Importantly, Table~\ref{table:table1} presents a framework updated on Adam optimizer~\citep{Kingma2014adam} to introduce one additional momentum term using a third-power gradient to improve the convergence bound. The \textit{HOME}-3 indicates that the highest power of the gradient for cultivating momentum is 3. Notably, the coordinate randomization $\mathcal{R}$ is only applied to nonsmooth nonconvex problems. Thus, the framework in Table \ref{table:table1} could be treated as a potential standard framework to incorporate high-order momentum.

 As discussed before, a higher-order momentum $S_t$ and $\hat{S}_t$ dominate the update at the beginning, due to $\left \|g_t^3 \right \| >> \left \|g_t \right \|$. Furthermore, when the gradient approximates a stationary point or local optimum, such as $\forall \epsilon >0, \left \|g_t \right \| < \epsilon$, the lower-power gradient is in charge of updating. In particular, let the Eq. \ref{eq3} equal to 0, we can infer the stopping criteria of \textit{HOME}-3:
 \begin{equation} \label{eq5}
\begin{gathered}
\forall \epsilon>0 \left \| \hat{M}_t - \hat{S}_t \right \| < \epsilon
\end{gathered}
\end{equation}
 Since $\left \| \hat{M}_t - \hat{S}_t \right \| < \epsilon$ can result in terminating \textit{HOME}-3, as indicated in \eqref{eq4} and \eqref{eq5}, we introduce coordinate randomization for \textit{HOME} optimizers to escape potential stationary points in the objective function. Furthermore, at the late stage, when the gradient approximates to the stationary point, such as $\left \|\hat{M}_t \right \|, \left \| \hat{S}_t \right \| < \epsilon$, coordinate randomization can maintain the difference between $\left \|\hat{M}_t \right \|$ and $\left \| \hat{S}_t \right \| $ in order to advance $\hat{S}_t - \hat{M}_t$ to escape an open cube of stationary points.

\section{Theoretical Analyses}

This section presents the convergence analyses of the \textit{HOME}-3 optimizer under three assumptions. We begin by examining the convex case that satisfies \textbf{Assumption}~\ref{assumption1}, demonstrating that \textit{HOME}-3 can achieve a convergence upper bound of $O(1 \slash T^{5 \slash 6})$, as outlined in Section 4.1. In Section 4.2, we extend this analysis under \textbf{Assumption}~\ref{assumption2}, showing that the convergence bound of the \textit{HOME}-3 optimizer remains comparable to that of the convex case. Additionally, in Section 4.3, we introduce a key advancement—coordinate randomization—which can further enhance the performance of \textit{HOME}-3 in nonsmooth nonconvex scenarios. The results partially answer the questions \textit{What is the role of randomization in dimension-free nonsmooth nonconvex optimization} raised by Jordan~\citep{jordan2023deterministic}. In short, complete theoretical proofs for the \textit{HOME}-3 optimizer are provided in Appendix A of the Supplementary Material.

\subsection{Convex Case}
We theoretically analyze the convergence bound of \textit{HOME}-3 under the convexity assumption (please refer to \textbf{Assumption} \ref{assumption1}) in this section.The following \textbf{Theorem} \ref{theorem1} demonstrates a convergence bound of \textit{HOME}-3 is $O(1 \slash T^{5 \slash 6})$.

\begin{theorem}
   \label{theorem1}
   Let $f$ satisfy \textbf{\textit{Assumption}}~\ref{assumption1}, suppose $T$ as the maximum iteration, according to \textbf{\textit{Definitions}}~\ref{def3},~\ref{def5}, and~\ref{def6}, then $\frac{\left \| \Sigma_{t=1}^T  (f(x_t)-f(x_T)) \right \|} {T} = O(1 \slash T^{5 \slash 6})$.
\end{theorem}

The detailed proof of \textbf{Theorem} \ref{theorem1} can be viewed in Appendix A, Supplementary Material.

\subsection{Smooth Nonconvex Case}
In this section, under the smooth nonconvex Assumption (please refer to \textbf{Assumption}~\ref{assumption2}, we prove that the convergence bound of \textit{HOME}-3 is $O(1 \slash T^{5 \slash 6})$. The potential issue impacting the convergence bound of \textit{HOME}-3 is the term $\frac{L}{2} \cdot \left \|x-y\right\|$. According to our analyses, if $T$ is sufficiently large and guarantees $\frac{L}{\sqrt{T}} \rightarrow 0, \forall x,y \in X$, in that case, the convergence bound of \textit{HOME}-3 is comparable to convexity assumption (please refer to \textbf{Assumption}~\ref{assumption1}). Similarly, the convergence upper bound of \textit{HOME}-3 under smooth nonconvex cases is $O(1 \slash T^{5 \slash 6})$.

\begin{theorem}
   \label{theorem2}
   Let $f$ satisfy \textbf{\textit{Assumption}}~\ref{assumption2}, suppose $T$ as the maximum iteration, according to \textbf{\textit{Definitions}}~\ref{def3},~\ref{def5}, and~\ref{def6}, then $\frac{\left \| f(x_t)-f(x_T) \right \|}{T} = O(1 \slash T^{5 \slash 6})$ holds.
\end{theorem}

The detailed proof of \textbf{Theorem}~\ref{theorem2} can be viewed in Appendix A, Supplementary Material.

\subsection{Nonsmooth Nonconvex Case}
Due to the complexity of nonsmooth nonconvex cases, $\left \| \hat{M}_t -\hat{S}_t \right \|$ could be 0 when the gradient approximates the stationary point. To overcome this challenge, we incorporate randomization to increase the opportunity for the optimizer to approximate an open cube of the global optimum. Notably, the following Lemma proves that the norm of coordinate randomization is equal to 1.

\begin{lemma}
\label{lemma1}
(Norm of Coordinate Randomization Operator is Equal to 1) Suppose the permutation randomization as an operator $\mathcal{R}:\mathbb{R}^D \rightarrow \mathbb{R}^D$, $\left \| \mathcal{R} \right \| =1$ holds, if $D < \infty$.
\end{lemma}

It is not difficult to prove \textbf{Lemma} \ref{lemma1}. The proof of Lemma \ref{lemma1} can be viewed in Appendix A, Supplementary Material.

Importantly, in \textbf{Theorem}~\ref{theorem3}, we discuss the upper bound on the convergence bound of gradient-based optimizer~\citep{wang2023parallelization} incorporating coordinate randomization is comparable to $\left \| \mathcal{G}^{t+1}  \cdot f(x_0) - \mathcal{G}^{t}  \cdot f(x_0)) \right \|$; thus, we discuss that coordinate randomization could maintain the convergence bound of incorporated gradient-based optimizer and is shown in \textbf{Theorem} \ref{theorem3}.

According to Definition \ref{def3}, we can infer:
\begin{equation} \label{eq6}
\begin{gathered}
\left \| \mathcal{R} \cdot \lbrace x_1, x_2, \cdots, x_D \rbrace 
 \right \|
 =  \left \| \lbrace \hat{x}_1, \hat{x}_2, \cdots, \hat{x}_D \rbrace \right \|
\end{gathered}
\end{equation}

According to \textbf{Definition}~\ref{def5}, \textbf{Lemma}~\ref{lemma1},for any $x, y \in I$, we have:
\begin{equation} \label{eq7}
\begin{gathered}
\left \| \mathcal{R}^t \cdot \mathcal{G}^t \cdot (f(x) -f(y)) \right \| \leq \left \| \mathcal{R}^t \right \| \cdot \left \| \mathcal{G}^t \cdot (f(x) -f(y)) \right \| =  \left \| \mathcal{G}^t  \cdot (f(x) -f(y)) \right \|
\end{gathered}
\end{equation}

Let $x$ be $x_1=\mathcal{G} \cdot f(x_0)$ and $Y$ be $x_0$, inferring from \eqref{eq5}, we have:
\begin{equation} \label{eq8}
\begin{gathered}
\left \| \mathcal{R}^t \cdot \mathcal{G}^t \cdot (f(x_1) -f(x_0)) \right \| \leq \  \left \| \mathcal{G}^{t+1}  \cdot f(x_0) - \mathcal{G}^{t}  \cdot f(x_0)) \right \|
\end{gathered}
\end{equation}

\begin{theorem}
\label{theorem3}
(Coordinate Randomization Maintains The Convergence Bound of Incorporated Optimizer) Inferring from \textbf{Lemma} \ref{lemma1}, the convergence bound of a gradient-based optimizer incorporating coordinate randomization $\mathcal{R} \cdot \mathcal{G}$ should be equal to the convergence bound of an original gradient-based optimizer $\mathcal{G}$ without coordinate randomization.
\end{theorem}

\section{Numerical Experiments}
We validate \textit{HOME} with three other peer optimizers, such as ADMM~\citep{Nishihara2015admm}, Adam \citep{Kingma2014adam}, and STORM \citep{cutkosky2019momentum}, on the public biomedical data in Multiband Multi-echo (MBME) functional Magnetic Resonance Imaging (fMRI)~\citep{MBMEfMRI}. After pre-processing \citep{ji2022empirical}, the size of each input signal matrix is $100 \times 902,629$. The total number of subjects is 29. In this empirical study, all optimizers are terminated after 100 iterations with other parameters fixed to the reported default values in the literature~\citep{Kingma2014adam, cutkosky2019momentum, Nishihara2015admm}. In addition, $\epsilon_2$ representing the difference between the previous and current gradient is the same as $\epsilon_1$~\citep{Kingma2014adam}. Furthermore, the experimental studies are validated on the CPU cluster, including 16 Intel Xeon X5570 2.93GHz. Moreover, to facilitate statistical analyses based on a large number of augmented subjects, the original 29 subjects are expanded to 100 via data augmentation techniques~\citep{wen2020time, iwana2021empirical}.

\subsection{Experiment on Convex Problem: Dictionary Learning}
Since Dictionary Learning (DL) is one of the representative alternative convex problems~\citep{hao2023dictionary, tovsic2011dictionary}, we employ \textit{HOME}-3 and other peer optimizers to optimize the objective functions of DL presented as follows:
\begin{equation} \label{eq9}
\begin{gathered}
\min_{X, Y \in \mathbb{R}^{p \times q}} \left \| I - XY \right \| + \lambda \left \| Y \right \|_1, p, q \in \mathbb{N}
\end{gathered} 
\end{equation}

In \eqref{eq9}, $I$ denotes the input matrix. $X$ and $Y$ denote weight and feature matrices, respectively. $\lambda$ represents a sparse trade-off set as the default value~\citep{tovsic2011dictionary}. Since DL is an alternative convex problem, we can validate the theoretical conclusion in Section 4.1. In addition, we provide a reconstruction loss to compare \textit{HOME} with other peer optimizers quantitatively. And, since DL is an unsupervised learning problem, we provide the reconstruction loss in Eq. \ref{eq10} as follows:
\begin{equation} \label{eq10}
\begin{gathered}
Reconstruction \; Loss = \frac{\left \| I - XY \right \|}{\left \| I \right \|}  
\end{gathered} 
\end{equation}

\begin{figure}[ht]
\begin{center}
\includegraphics[width=0.86\textwidth]{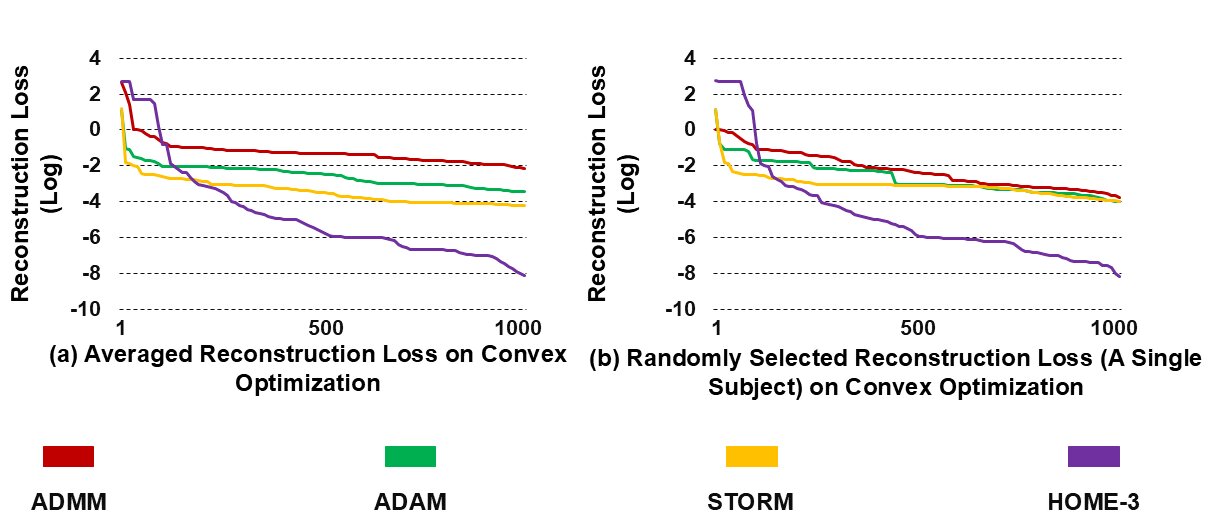}
\end{center}
\caption{Averaged reconstruction loss comparison of proposed \textit{HOME}-3 and other three peer optimizers within one hundred iterations}\label{fig:fig1}
\end{figure}

Overall, Figure \ref{fig:fig1} presents the averaged reconstruction loss of \textit{HOME}-3 and other peer optimizers to optimize the objective function of DL. In particular, according to Figure \ref{fig:fig1} (a), \textit{HOME}-3 can enhance the convergence and reconstruction accuracy. Notably, \textit{HOME}-3 demonstrates a more extensive reconstruction loss at the early stage due to a larger norm of high-power gradient. In Figure \ref{fig:fig1} (b), in this most straightforward case, an individual reconstruction loss reveals the convergence of ADMM~\citep{Nishihara2015admm} is faster than Adam~\citep{Kingma2014adam} and STORM~\citep{cutkosky2019momentum} but \textit{HOME}-3 obtains the steepest convergence curve at the early stage.

\subsection{Experiment on Smooth Nonconvex Problem: Deep Nonlinear Matrix Factorizations}
Furthermore, to validate \textit{HOME}-3 on smooth nonconvex optimization, we introduce the objective functions of Deep Nonlinear Matrix Factorization (DNMF)~\citep{trigeorgis2016deep}, presented in \eqref{eq11a} and \eqref{eq11b}. Overall, DNMF is comparable to layer-stack deep neural networks such as a Deep Belief Network (DBN) consisting of multiple restricted Boltzmann machines~\citep{hinton2009deep, gu2022approximation}. Meanwhile, similar to DBN, since DNMF is an unsupervised learning problem, we focus on comparing reconstruction loss in the following Figure \ref{fig:fig2}. Importantly, to avoid arbitrary hyperparameter tuning, we employ a rank estimator \citep{zhao2020low} to automatically estimate the number of layers and layer size. For activation function between adjacent layers, considering previous works \citep{jordan2023deterministic}, we set Rectified Linear Unit (ReLU)~\citep{agarap2018deep} as an activation function $\mathcal{N}_k$ in \eqref{eq11b} to increase the complexity of objective function in DNMF.
\begin{subequations} \label{eq11}
\begin{align}
\min_{Z_i \in \mathbb{R}^{p \times q}} \bigcup_{i=1}^k \left \| Z_i\right \|_1 \label{eq11a}\\
\textit{s.t.} (\prod_{i=1}^k X_i) \cdot \mathcal{N}_k (Y_k) + Z_k=I \label{eq11b}
\end{align} 
\end{subequations}
In \eqref{eq11}, $I$ denotes the input matrix. $X_i$ denotes the current layer and $Y_i$ denotes the current feature matrix. In addition, $\mathcal{N}_k$ represents an activation function in the current layer. Lastly, $Z_k$ indicates a background noise matrix. And $k$ represents the total layer number.

In addition, reconstruction loss under smooth nonconvex assumption is denoted as:
\begin{equation} \label{eq12}
\begin{gathered}
Reconstruction \; Loss = \frac{\left \| (\prod_{i=1}^k X_i) \cdot \mathcal{N}_k (Y_k) + Z_k - I \right \|}{\left \| I \right \|}  
\end{gathered} 
\end{equation}
In the following Figure \ref{fig:fig2}, we present a reconstruction loss to compare the \textit{HOME}-3 with other peer optimizers in the first and second layers of DNMF. Overall, in Figure \ref{fig:fig2} (a) and (b), \textit{HOME}-3 has improved the convergence. Even in the late stage (after 60 iterations), due to the high-order momentum, \textit{HOME}-3 can still converge faster than peer optimizers.

\begin{figure}[ht]
   \begin{center}
    \includegraphics[width=0.86\textwidth]{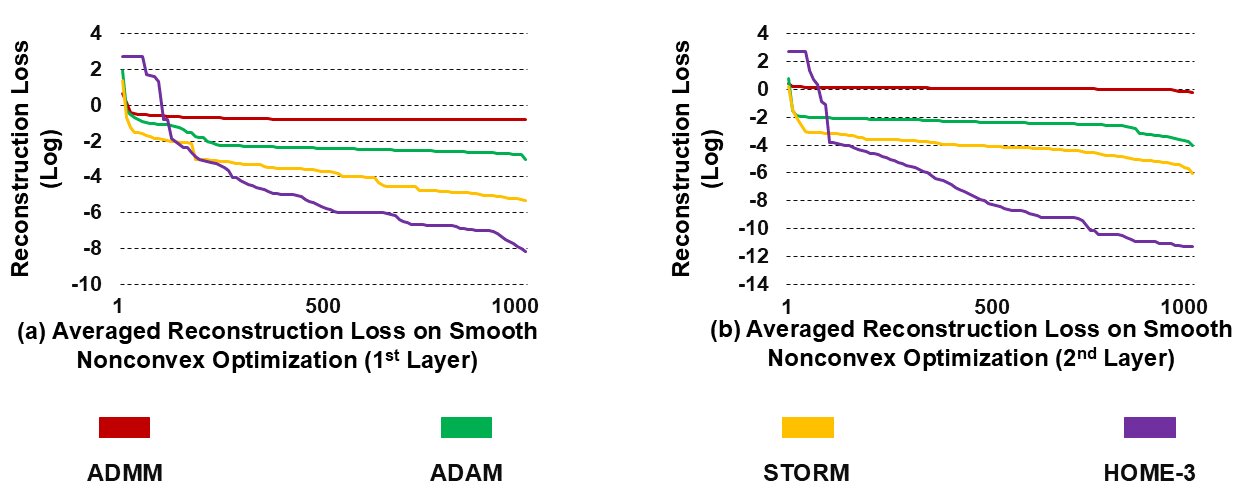}
    \caption{Averaged reconstruction loss comparison of proposed \textit{HOME}-3 and other three peer optimizers with
    in one hundred iterations at first and second layers of DNMF}\label{fig:fig2}
    \end{center}
\end{figure}

\subsection{Experiment on Nonsmooth Nonconvex Problem: Noisy Deep Matrix Factorization}

Moreover, in this section, to continuously increase the complexity in objective functions, we aim to investigate the performance of \textit{HOME}-3 optimizer under the nonsmooth nonconvex case. To implement a nonsmooth nonconvex optimization, we add additional random noise to the feature matrix in DNMF~\citep{lu2014generalized, lin2022gradient}, such as:
\begin{equation} \label{eq13}
\begin{gathered}
Y_i \leftarrow Y_i + random \; noise 
\end{gathered} 
\end{equation}

In~\eqref{eq13}, a random noise is added to the feature matrix $Y_i$ in \eqref{eq11}. The random noise results in nonsmooth nonconvex objective functions~\citep{lu2014generalized, lin2022gradient}. Importantly, to avoid the noise overwhelming the original data, we set the boundary of random noise in this experiment as $\lbrack -0.1 \cdot Median, 0.1 \cdot Median \rbrack$. $Median$ represents the median of the input matrix or vector.

\begin{figure}[H]
    \begin{center}
    \includegraphics[width=0.86\textwidth]{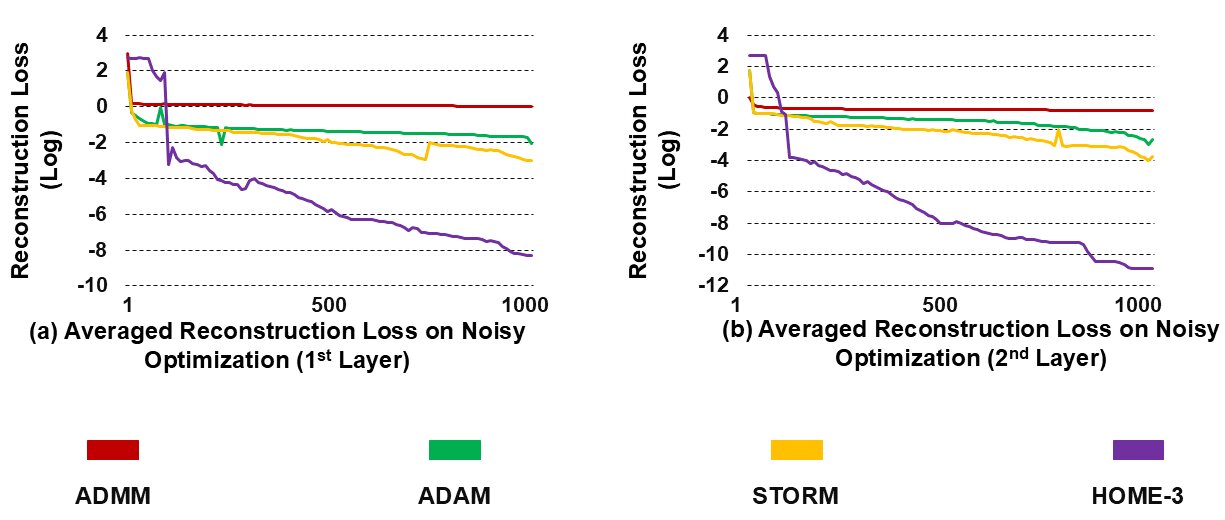}
    \caption{The averaged training loss comparison of proposed \textit{HOME}-3 and other three peer optimizers within one hundred iterations of all subjects at first and second layers of noisy DNMF, respectively.}\label{fig:fig3}
    \end{center}
\end{figure}

Figure~\ref{fig:fig3} compares reconstruction loss of \textit{HOME}-3 with other peer optimizers under the nonsmooth nonconvex case. Even in the most complex case, \textit{HOME}-3 can still enhance the convergence and provide most accurate reconstruction. In Figures \ref{fig:fig3} (a) and (b), it is noticeable that the convergence curve of \textit{HOME}-3 is steepest within 200 iterations. The results further demonstrate that the high-order momentum can improve the convergence and maintain the impact until the late stage. Importantly, additional experimrential results using DNN and logistic regression can be viewed in Figures~\ref{fig:fig6} and~\ref{fig:fig5}.

\subsection{Statistical Analyses}
\begin{figure}[ht]
    \begin{center}
    \includegraphics[width=0.80\textwidth]{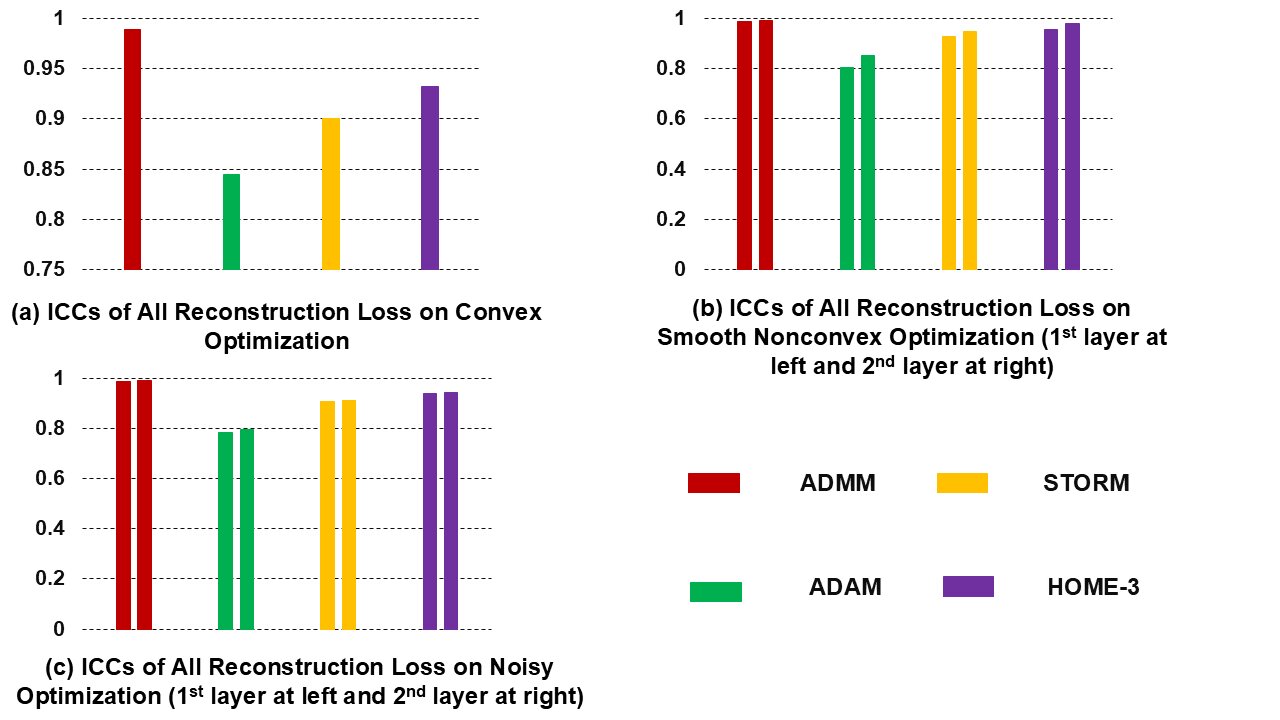}
    \caption{Consistency and robustness comparisons of the proposed \textit{HOME}-3 and three peer algorithms are presented.}
    \label{fig:fig4}
    \end{center}
\end{figure}

In this section, we quantitatively analyze previous experimental results on a large number of samples. Notably, the non-independency limits of iterative optimizer to directly employ a \textit{t-test} and$\slash$or confidential intervals to compare all iterative reconstruction accuracy is not suitable~\citep{field2013discovering}. Alternatively, Intra-class correlation coefficients (ICCs), a descriptive statistic technique that can be used for quantitative measurements organized into groups~\citep{bujang2017simplified}. In Figures~\ref{fig:fig4} (a), (b), and (c), we report the ICCs of \textit{HOME}-3 and three other peer optimizers on previous empirical experiments in Sections 5.1, 5.2, and 5.3. In particular, Figure \ref{fig:fig4} (a) describes the ICCs on reconstruction loss of \textit{HOME}-3, ADMM~\citep{Nishihara2015admm}, Adam~\citep{Kingma2014adam}, and STORM~\citep{cutkosky2019momentum} on 100 subjects. ADMM is the most robust on convex optimization, and \textit{HOME}-3 is more robust than Adam and STORM~\citep{Kingma2014adam, cutkosky2019momentum}. In addition, Figure \ref{fig:fig4} (b) presents the robustness of \textit{HOME}-3, ADMM~\citep{Nishihara2015admm}, Adam~\citep{Kingma2014adam}, and STORM~\citep{cutkosky2019momentum} on smooth nonconvex optimization using 100 subjects. In particular, \textit{HOME}-3 achieves the most robust reconstruction accuracy since the ICCs in both the first and second layers are close to 0.93 and 0.95. Although ADMM obtains the largest ICCs, its reconstruction loss is inaccurate in Figure \ref{fig:fig2}. Notably, though coordinate randomization is introduced, \textit{HOME}-3 is more consistent than Adam and STORM on smooth nonconvex optimization. Lastly, in Figure \ref{fig:fig4} (c), the robustness of \textit{HOME}-3 is higher than Adam and STORM. There is no significant difference between the first and second layers using \textit{HOME}-3 to optimize nonsmooth nonconvex deep models.

\section{Conclusion}
This work introduces an innovative high-order momentum technique that utilizes high-power gradients to significantly enhance the performance of the gradient-based optimizer. Our contributions are both theoretical and empirical. On the theoretical side, we demonstrate that high-order momentum improves the convergence bound of optimizers in both convex and smooth nonconvex cases, achieving an upper bound of $O(1 \slash T^{5 \slash 6})$. Empirically, extensive experiments showcase that \textit{HOME}-3 consistently delivers superior reconstruction accuracy across convex, smooth nonconvex, and nonsmooth nonconvex problems, underscoring its robustness. Looking ahead, an exciting direction for future research is determining the optimal order of momentum for complex objective functions, which will be pivotal in efficiently optimizing Large Language Models.
%\subsubsection*{Author Contributions}
%If you'd like to, you may include  a section for author contributions as is done
%in many journals. This is optional and at the discretion of the authors.

\bibliography{home3_bib}

\begin{thebibliography}{50}
\providecommand{\natexlab}[1]{#1}
\providecommand{\url}[1]{\texttt{#1}}
\expandafter\ifx\csname urlstyle\endcsname\relax
  \providecommand{\doi}[1]{doi: #1}\else
  \providecommand{\doi}{doi: \begingroup \urlstyle{rm}\Url}\fi

\bibitem[Agarap(2018)]{agarap2018deep}
A.~F. Agarap.
\newblock Deep learning using rectified linear units (relu).
\newblock \emph{arXiv preprint arXiv:1803.08375}, 2018.

\bibitem[Allen-Zhu(2017)]{Allen2017firstdirect}
Z.~Allen-Zhu.
\newblock The first direct acceleration of stochastic gradient methods.
\newblock \emph{Journal of Machine Learning Research}, 18\penalty0 (1):\penalty0 8194--8244, 2017.

\bibitem[Arjevani and Shamir(2015)]{Arjevani2015complexityl}
Y.~Arjevani and O.~Shamir.
\newblock Communication complexity of distributed convex learning and optimization.
\newblock In \emph{Advances in neural information processing systems}, volume~28, 2015.

\bibitem[Beznosikov et~al.(2023)Beznosikov, Gorbunov, Berard, and Loizou]{beznosikov2023stochastic}
A.~Beznosikov, E.~Gorbunov, H.~Berard, and N.~Loizou.
\newblock Stochastic gradient descent-ascent: Unified theory and new efficient methods.
\newblock In \emph{International Conference on Artificial Intelligence and Statistics}, pages 172--235. PMLR, 2023.

\bibitem[Bujang and Baharum(2017)]{bujang2017simplified}
M.~A. Bujang and N.~Baharum.
\newblock A simplified guide to determination of sample size requirements for estimating the value of intraclass correlation coefficient: a review.
\newblock \emph{Archives of Orofacial Science}, 12\penalty0 (1), 2017.

\bibitem[Chandra et~al.(2022)Chandra, Xie, Ragan-Kelley, and Meijer]{chandra2022gradient}
K.~Chandra, A.~Xie, J.~Ragan-Kelley, and E.~Meijer.
\newblock Gradient descent: The ultimate optimizer.
\newblock \emph{Advances in Neural Information Processing Systems}, 35:\penalty0 8214--8225, 2022.

\bibitem[Clarke(1974)]{clarke1974necessary}
F.~H. Clarke.
\newblock Necessary conditions for nonsmooth variational problems.
\newblock In \emph{Optimal Control Theory and its Applications: Proceedings of the Fourteenth Biennial Seminar of the Canadian Mathematical Congress University of Western Ontario, August 12--25, 1973}, pages 70--91. Springer, 1974.

\bibitem[Clarke(1975)]{clarke1975generalized}
F.~H. Clarke.
\newblock Generalized gradients and applications.
\newblock \emph{Transactions of the American Mathematical Society}, 205:\penalty0 247--262, 1975.

\bibitem[Clarke(1981)]{clarke1981generalized}
F.~H. Clarke.
\newblock Generalized gradients of lipschitz functionals.
\newblock \emph{Advances in Mathematics}, 40\penalty0 (1):\penalty0 52--67, 1981.

\bibitem[Clarke(1990)]{clarke1990optimization}
F.~H. Clarke.
\newblock \emph{Optimization and nonsmooth analysis}.
\newblock SIAM, 1990.

\bibitem[Cutkosky and Orabona(2019)]{cutkosky2019momentum}
A.~Cutkosky and F.~Orabona.
\newblock Momentum-based variance reduction in non-convex sgd.
\newblock \emph{Advances in neural information processing systems}, 32, 2019.

\bibitem[Defazio et~al.(2014)Defazio, Bach, and Lacoste-Julien]{Defazio2014saga}
A.~Defazio, F.~Bach, and S.~Lacoste-Julien.
\newblock Saga: A fast incremental gradient method with support for non-strongly convex composite objectives.
\newblock In \emph{Advances in neural information processing systems}, volume~27, 2014.

\bibitem[Field(2013)]{field2013discovering}
A.~Field.
\newblock \emph{Discovering statistics using IBM SPSS statistics}.
\newblock sage, 2013.

\bibitem[Gu et~al.(2022)Gu, Yang, and Zhou]{gu2022approximation}
L.~Gu, L.~Yang, and F.~Zhou.
\newblock Approximation properties of gaussian-binary restricted boltzmann machines and gaussian-binary deep belief networks.
\newblock \emph{Neural Networks}, 153:\penalty0 49--63, 2022.

\bibitem[Haji and Abdulazeez(2021)]{haji2021comparison}
S.~H. Haji and A.~M. Abdulazeez.
\newblock Comparison of optimization techniques based on gradient descent algorithm: A review.
\newblock \emph{PalArch's Journal of Archaeology of Egypt/Egyptology}, 18\penalty0 (4):\penalty0 2715--2743, 2021.

\bibitem[Hao et~al.(2023)Hao, Stuart, Kowalski, Choudhary, Hoffman, Hartman, Srivastava, Molla, Madad, Fernandez-Granda, et~al.]{hao2023dictionary}
Y.~Hao, T.~Stuart, M.~H. Kowalski, S.~Choudhary, P.~Hoffman, A.~Hartman, A.~Srivastava, G.~Molla, S.~Madad, C.~Fernandez-Granda, et~al.
\newblock Dictionary learning for integrative, multimodal and scalable single-cell analysis.
\newblock \emph{Nature Biotechnology}, pages 1--12, 2023.

\bibitem[Hazan et~al.(2007)Hazan, Agarwal, and Kale]{Hazan2007logarithmic}
E.~Hazan, A.~Agarwal, and S.~Kale.
\newblock Logarithmic regret algorithms for online convex optimization.machine learning.
\newblock \emph{Machine Learning}, 69\penalty0 (2):\penalty0 169--192, 2007.

\bibitem[Hinton(2009)]{hinton2009deep}
G.~E. Hinton.
\newblock Deep belief networks.
\newblock \emph{Scholarpedia}, 4\penalty0 (5):\penalty0 5947, 2009.

\bibitem[Huang et~al.(2021)Huang, Li, and Huang]{huang2021super}
F.~Huang, J.~Li, and H.~Huang.
\newblock Super-adam: faster and universal framework of adaptive gradients.
\newblock \emph{Advances in Neural Information Processing Systems}, 34:\penalty0 9074--9085, 2021.

\bibitem[Iwana and Uchida(2021)]{iwana2021empirical}
B.~K. Iwana and S.~Uchida.
\newblock An empirical survey of data augmentation for time series classification with neural networks.
\newblock \emph{Plos one}, 16\penalty0 (7):\penalty0 e0254841, 2021.

\bibitem[Ji et~al.(2022)Ji, Hendrix, and Thomason]{ji2022empirical}
L.~Ji, C.~L. Hendrix, and M.~E. Thomason.
\newblock Empirical evaluation of human fetal fmri preprocessing steps.
\newblock \emph{Network Neuroscience}, 6\penalty0 (3):\penalty0 702--721, 2022.

\bibitem[Johnson and Zhang(2013)]{Johnson2013accelerating}
R.~Johnson and T.~Zhang.
\newblock Accelerating stochastic gradient descent using predictive variance reduction.
\newblock In \emph{Advances in neural information processing systems}, volume~26, 2013.

\bibitem[Jordan et~al.(2023)Jordan, Kornowski, Lin, Shamir, and Zampetakis]{jordan2023deterministic}
M.~Jordan, G.~Kornowski, T.~Lin, O.~Shamir, and M.~Zampetakis.
\newblock Deterministic nonsmooth nonconvex optimization.
\newblock In \emph{The Thirty Sixth Annual Conference on Learning Theory}, pages 4570--4597. PMLR, 2023.

\bibitem[Kingma and Ba(2014)]{Kingma2014adam}
D.~P. Kingma and J.~Ba.
\newblock Adam: A method for stochastic optimization.
\newblock \emph{arXiv:1412.6980}, 2014.

\bibitem[Levy et~al.(2021)Levy, Kavis, and Cevher]{levy2021storm+}
K.~Y. Levy, A.~Kavis, and V.~Cevher.
\newblock Storm+: Fully adaptive sgd with momentum for nonconvex optimization.
\newblock \emph{arXiv preprint arXiv:2111.01040}, 2021.

\bibitem[Lin et~al.(2015)Lin, Mairal, and Harchaoui]{Lin2015universal}
H.~Lin, J.~Mairal, and Z.~Harchaoui.
\newblock A universal catalyst for first-order optimization.
\newblock In \emph{Advances in neural information processing systems}, volume~28, 2015.

\bibitem[Lin et~al.(2022)Lin, Zheng, and Jordan]{lin2022gradient}
T.~Lin, Z.~Zheng, and M.~Jordan.
\newblock Gradient-free methods for deterministic and stochastic nonsmooth nonconvex optimization.
\newblock \emph{Advances in Neural Information Processing Systems}, 35:\penalty0 26160--26175, 2022.

\bibitem[Liu et~al.(2020)Liu, Gao, and Yin]{liu2020improved}
Y.~Liu, Y.~Gao, and W.~Yin.
\newblock An improved analysis of stochastic gradient descent with momentum.
\newblock \emph{Advances in Neural Information Processing Systems}, 33:\penalty0 18261--18271, 2020.

\bibitem[Loizou and Richt{\'a}rik(2020)]{loizou2020momentum}
N.~Loizou and P.~Richt{\'a}rik.
\newblock Momentum and stochastic momentum for stochastic gradient, newton, proximal point and subspace descent methods.
\newblock \emph{Computational Optimization and Applications}, 77\penalty0 (3):\penalty0 653--710, 2020.

\bibitem[Lu et~al.(2014)Lu, Tang, Yan, and Lin]{lu2014generalized}
C.~Lu, J.~Tang, S.~Yan, and Z.~Lin.
\newblock Generalized nonconvex nonsmooth low-rank minimization.
\newblock In \emph{Proceedings of the IEEE conference on computer vision and pattern recognition}, pages 4130--4137, 2014.

\bibitem[Lydia and Francis(2019)]{lydia2019adagrad}
A.~Lydia and S.~Francis.
\newblock Adagrad—an optimizer for stochastic gradient descent.
\newblock \emph{Int. J. Inf. Comput. Sci}, 6\penalty0 (5):\penalty0 566--568, 2019.

\bibitem[Mai and Johansson(2020)]{mai2020convergence}
V.~Mai and M.~Johansson.
\newblock Convergence of a stochastic gradient method with momentum for non-smooth non-convex optimization.
\newblock In \emph{International conference on machine learning}, pages 6630--6639. PMLR, 2020.

\bibitem[Nemirovski et~al.(2009)Nemirovski, Juditsky, Lan, and Shapiro]{Nemirovski2009robuststochastic}
A.~Nemirovski, A.~Juditsky, G.~Lan, and A.~Shapiro.
\newblock Robust stochastic approximation approach to stochastic programming.
\newblock \emph{SIAM Journal on optimization}, 19\penalty0 (4):\penalty0 1574--1609, 2009.

\bibitem[Nishihara et~al.(2015)Nishihara, Lessard, Recht, Packard, and Jordan]{Nishihara2015admm}
R.~Nishihara, L.~Lessard, B.~Recht, A.~Packard, and M.~Jordan.
\newblock A general analysis of the convergence of admm.
\newblock In \emph{International Conference on Machine Learning}, pages 343--352, 2015.

\bibitem[Rakhlin et~al.(2011)Rakhlin, Shamir, and Sridharan]{Rakhlin2011stronglyconvex}
A.~Rakhlin, O.~Shamir, and K.~Sridharan.
\newblock Making gradient descent optimal for strongly convex stochastic optimization.
\newblock \emph{arXiv preprint arXiv:1109.5647}, 2011.

\bibitem[Reddi et~al.(2019)Reddi, Kale, and Kumar]{reddi2019convergence}
S.~J. Reddi, S.~Kale, and S.~Kumar.
\newblock On the convergence of adam and beyond.
\newblock \emph{arXiv preprint arXiv:1904.09237}, 2019.

\bibitem[Rudin(1973)]{Rudin1973functionalanalysis}
W.~Rudin.
\newblock \emph{Functional analysis.}
\newblock McGraw-Hill, University of Michigan, 2 edition, 1973.

\bibitem[Schober and Vetter(2021)]{schober2021logistic}
P.~Schober and T.~R. Vetter.
\newblock Logistic regression in medical research.
\newblock \emph{Anesthesia \& Analgesia}, 132\penalty0 (2):\penalty0 365--366, 2021.

\bibitem[Shalev-Shwartz and Zhang(2013)]{shalev2013stochasticdual}
S.~Shalev-Shwartz and T.~Zhang.
\newblock Stochastic dual coordinate ascent methods for regularized loss minimization.
\newblock \emph{Journal of Machine Learning Research}, 14\penalty0 (2), 2013.

\bibitem[Shut(2023)]{BreastCancer}
M.~Shut.
\newblock Breast cancer data.
\newblock \url{https://www.kaggle.com/datasets/marshuu/breast-cancer?resource=download}, 2023.
\newblock Accessed: 2023-01-01.

\bibitem[To{\v{s}}i{\'c} and Frossard(2011)]{tovsic2011dictionary}
I.~To{\v{s}}i{\'c} and P.~Frossard.
\newblock Dictionary learning.
\newblock \emph{IEEE Signal Processing Magazine}, 28\penalty0 (2):\penalty0 27--38, 2011.

\bibitem[Trigeorgis et~al.(2016)Trigeorgis, Bousmalis, Zafeiriou, and Schuller]{trigeorgis2016deep}
G.~Trigeorgis, K.~Bousmalis, S.~Zafeiriou, and B.~W. Schuller.
\newblock A deep matrix factorization method for learning attribute representations.
\newblock \emph{IEEE transactions on pattern analysis and machine intelligence}, 39\penalty0 (3):\penalty0 417--429, 2016.

\bibitem[Wang(2018)]{MBMEfMRI}
A.~D. C. S. N. R. M. L.~Y. Wang.
\newblock Multiband multi-echo bold fmri.
\newblock \url{https://openneuro.org/datasets/ds000216/versions/00001}, 2018.
\newblock Accessed: 2018-07-17.

\bibitem[Wang and Shen(2023)]{wang2023parallelization}
L.~Wang and B.~Shen.
\newblock On the parallelization upper bound for asynchronous stochastic gradients descent in non-convex optimization.
\newblock \emph{Journal of Optimization Theory and Applications}, 196\penalty0 (3):\penalty0 900--935, 2023.

\bibitem[Wang and Wen(2022)]{wang2022proximal}
Z.~Wang and B.~Wen.
\newblock Proximal stochastic recursive momentum algorithm for nonsmooth nonconvex optimization problems.
\newblock \emph{Optimization}, pages 1--15, 2022.

\bibitem[Wang et~al.(2021)Wang, Zhang, Chang, Li, and Luo]{wang2021distributed}
Z.~Wang, J.~Zhang, T.-H. Chang, J.~Li, and Z.-Q. Luo.
\newblock Distributed stochastic consensus optimization with momentum for nonconvex nonsmooth problems.
\newblock \emph{IEEE Transactions on Signal Processing}, 69:\penalty0 4486--4501, 2021.

\bibitem[Wen et~al.(2020)Wen, Sun, Yang, Song, Gao, Wang, and Xu]{wen2020time}
Q.~Wen, L.~Sun, F.~Yang, X.~Song, J.~Gao, X.~Wang, and H.~Xu.
\newblock Time series data augmentation for deep learning: A survey.
\newblock \emph{arXiv preprint arXiv:2002.12478}, 2020.

\bibitem[Zhang and Bao(2022)]{zhang2022sadam}
W.~Zhang and Y.~Bao.
\newblock Sadam: Stochastic adam, a stochastic operator for first-order gradient-based optimizer.
\newblock \emph{arXiv preprint arXiv:2205.10247}, 2022.

\bibitem[Zhang et~al.(2012)Zhang, Wainwright, and Duchi]{Zhang2012cimmuncation}
Y.~Zhang, M.~J. Wainwright, and J.~C. Duchi.
\newblock Communication-efficient algorithms for statistical optimization.
\newblock In \emph{Advances in neural information processing systems}, volume~25, 2012.

\bibitem[Zhao and Zhao(2020)]{zhao2020low}
J.~Zhao and L.~Zhao.
\newblock Low-rank and sparse matrices fitting algorithm for low-rank representation.
\newblock \emph{Computers \& Mathematics with Applications}, 79\penalty0 (2):\penalty0 407--425, 2020.

\end{thebibliography}
%%%%%%%%%%%%%%%%%%%%%%%%%%%%%%%%%%%%%%%%%%%%%%%%%%%%%%%%%%%%%%%%%%%%%%%%%%%%%%%%%%%%%%%%%%%%%%%%%%%%%%
\newpage
\appendix
\section{Appendix}

The definitions and explanations of all mathematical symbols are illustrated in Table~\ref{table:table3}
\begin{table}[H]
\centering
\caption{The definitions of mathematical symbols}
\label{tab:symbols}
\begin{tabular}{>{\centering\arraybackslash}m{4cm} >{\centering\arraybackslash}m{10cm}}
\toprule
\textbf{Symbol} & \textbf{Description} \\
\midrule
\( f(x) \) & Objective function \\
\( x_t \in \mathbb{R}^D \) &  Variable in a single dimension of Objective function at iteration \( t \) \\
\( g_t = \nabla f(x_t) \) & Gradient at iteration \( t \) \\
\( g_t^n \) &  $n^{th}$ power of a gradient \\
\( M_t \) & First momentum term at iteration \( t \) \\
\( V_t \) & Second moment term (squared gradients) \\
\( S_t \) & Third moment term (cubed gradients) \\
\(\alpha_t \) & Learning rate at $t$ iteration \\
\( \mathcal{G} \) & Gradient operator \\
\( \mathcal{R} \) & Coordinate randomization operator \\
\( \hat{x}_t \) & Output after applying randomization \\
\( D \) & Dimension of the input space \\
\( T \) & Total number of iterations \\
\( \epsilon, \epsilon_1, \epsilon_2 \) & Convergence thresholds \\
\bottomrule
\end{tabular}
\label{table:table3}
\end{table}

We present the pseudocode of \textit{HOME}-3 in Table~\ref{table:table1}:
\begin{table}[H]
  \caption{The Pseudocode of High-Order Momentum Estimator (HOME-3)}
  \centering
  \begin{tabular}{l}
    \toprule
   \textbf{Algorithm 1:} \textit{HOME}-3\\
    \midrule
    1:\,\, \textbf{while} $t < T$ \textbf{do} \\
    2:\,\,\,\, $g_t \leftarrow \nabla_x f(x_t)$ \hfill \textit{: Compute gradient} \\
    3:\,\,\,\, $M_t \leftarrow \beta_1 M_{t-1} + (1-\beta_1)g_t$ \hfill \textit{: First moment} \\
    4:\,\,\,\, $V_t \leftarrow \beta_2 V_{t-1} + (1-\beta_2)g_t^2$ \hfill \textit{: Second moment} \\
    5:\,\,\,\, $S_t \leftarrow \beta_3 S_{t-1} + (1-\beta_3)g_t^3$ \hfill \textit{: Third moment} \\
    6:\,\,\,\, $\hat{M}_t \leftarrow \frac{M_t}{1-\beta_1^t}$ \\
    7:\,\,\,\, $\hat{V}_t \leftarrow \frac{V_t}{1-\beta_2^t}$ \\
    8:\,\,\,\, $\hat{S}_t \leftarrow \frac{S_t}{1-\beta_3^t}$ \\
    9:\,\,\,\, $x_{t+1} \leftarrow x_t - \alpha_t \cdot (\hat{M}_t - \hat{S}_t) / (\sqrt{\hat{V}_t}+\epsilon_1)$ \hfill \textit{: Update rule} \\
    10:\,\, \textbf{if} $ \| \hat{M}_t - \hat{S}_t \| < \epsilon_2 $ \textbf{then} \hfill \textit{: Check for stationarity} \\
    11:\,\,\,\,\,\, $\hat{x}_{t+1} \leftarrow \mathcal{R}(x_{t+1})$ \hfill \textit{: Apply randomization} \\
    12:\,\,\,\,\,\, $x_{t+1} \leftarrow \hat{x}_{t+1}$ \\
    13:\,\, \textbf{end if} \\
    14:\,\, $t \leftarrow t + 1$ \\
    15:\,\, \textbf{end while} \\
    \bottomrule
  \end{tabular}
  \label{table:table1}
\end{table}

\textit{\textbf{Proofs of Theorems}}:

\textbf{Theorem 4.1} Let $f$ satisfy \textbf{\textit{Assumption}} 1, suppose $T$ as the maximum iteration, inferring from \textbf{\textit{Definitions}} 3, 5, and 6, then $\frac{\left \|\Sigma_{t=1}^T  (f(x_t)-f(x_T)) \right \|}{T} = O(1 \slash T^{5 \slash 6})$ holds.

\noindent \textbf{\textit{Proof}}: According to \textbf{Theorem} 10.5 in Kingma's work~\cite{Kingma2014adam} and Theorem 4 in Reddi's work~\cite{reddi2019convergence} , suppose the current iteration is $t$, we have the iterative format of \textit{HOME}-3 as:
\begin{equation} \tag{A1} \label{eqA1}
\begin{gathered}
x_{t+1} = x_t - \alpha \cdot \frac{\hat{M}_t - \hat{S}_t}{\sqrt{\hat{V}_t}}
\end{gathered}
\end{equation}

Then, we subtract scalar $x_T$ and square the both side of \eqref{eqA1},
\begin{equation} \tag{A2} \label{eqA2}
\begin{gathered}
(x_{t+1} - x_T)^2 = (x_t - x_T)^2 - 2 \alpha \cdot \frac{(\hat{M}_t - \hat{S}_t)}{\sqrt{\hat{V}_t}} \cdot (x_t - x_T) + \alpha^2 \cdot (\frac{\hat{M}_t - \hat{S}_t}{\sqrt{\hat{V}_t}})^2
\end{gathered}
\end{equation}

Inferring from \eqref{eqA2}, due to initial value $\hat{S}_0$ equal to 0, $\hat{S}_t$ can be considered a linear combination of cubed gradient $g_t^3$:
\begin{equation} \tag{A3} \label{eqA3}
\begin{gathered}
\hat{S}_t = k_1 \cdot g_1^3 + k_2 \cdot g_2^3 + \cdots + k_t \cdot g_t^3
\end{gathered}
\end{equation}
In \eqref{eqA3}, $\lbrace k_i \rbrace_{i=1}^t$ is coefficient only relating to $\beta_3$.

Next, inferring from \textbf{Definition} \ref{def3}, $\hat{S}_t$ is bounded. We have:
\begin{equation} \tag{A4} \label{eqA4}
\begin{gathered}
\left \| \hat{S}_t \right \| \leq   max (\left \|\lbrace k_i \rbrace_{i=1}^t \right \|) \cdot max (\left \| \lbrace g_t \rbrace_{t=1}^T\right \|) 
\end{gathered}
\end{equation}

Similarly, inferring from \eqref{eqA4}, we can prove that the first and second momentum, $\hat{M}_t$ and $\hat{V}_t$, are also bounded. Hereby, according to \eqref{eqA4}, we categorize the convergence bound under convexity into two folds:

1). When $g_t$ is sufficiently large, for example $\left \| g_t \right \| > 1$, we have $\left \| g_t^3 \right \| >> \left \| g_t \right \|$. Thus, when $g_t$ is sufficiently large to conveniently analyze the convergence bound, we can ignore the influence from $\hat{M}_t$. In that case, inferring from \eqref{eqA4}, we have:
\begin{equation} \tag{A5} \label{eqA5}
\begin{gathered}
(x_{t+1}-x_T)^2 = (x_t-x_T)^2 +2\frac{\alpha}{\sqrt{\hat{V}_t}}(\beta_3 S_{t-1}+(1-\beta_3)g_t^3) (x_t-x_T) +\alpha^2\frac{\hat{S}^2}{\hat{V}_t}
\end{gathered}
\end{equation}

We can infer from \eqref{eqA5}:
\begin{equation} \tag{A6} \label{eqA6}
\begin{gathered}
g_t^3(x_T-x_t) = \frac{\sqrt{\hat{V}_t}}{2\alpha_t(1-\beta_3)}[(x_t-x_T)^2 - (x_{t+1}-x_T)^2 ]+ \frac{\beta_3}{1-\beta_3}S_{t-1}+\frac{\alpha_t}{1-\beta_3} \cdot \frac{\hat{S}^2}{\sqrt{\hat{V}_t}}
\end{gathered}
\end{equation}

The \eqref{eqA6} can be converted to the following:
\begin{equation} \tag{A7} \label{eqA7}
\begin{gathered}
g_t^3(x_T-x_t) = \frac{\sqrt{\hat{V}_t}}{2\alpha(1-\beta_3)}\lbrack (x_t-x_T)^2 - (x_{t+1}-x_T)^2 \rbrack + \\
\frac{\beta_3}{1-\beta_3} \frac{\hat{V}_t^{\frac{1}{4}}}{\sqrt{\alpha}} \frac{\sqrt{\alpha} S_{t-1}}{\hat{V}_t^{\frac{1}{4}}} (x_t-x_T) +
\frac{\alpha}{1-\beta_3} \cdot \frac{\hat{S}^2}{\sqrt{\hat{V}_t}} 
\end{gathered}
\end{equation}

Using Young's inequality ($ab \leq \frac{1}{2}(a^2 + b^2)$), we can infer:
\begin{equation} \tag{A8} \label{eqA8}
\begin{gathered}
g_t^3(x_T-x_t) \leq \frac{\sqrt{\hat{V}_t}}{2\alpha(1-\beta_3)}\lbrack (x_t-x_T)^2 - (x_{t+1}-x_T)^2 \rbrack + \\
\frac{\beta_3}{2\alpha(1-\beta_3)} (x_t-x_T)^2 \sqrt{\hat{V}_{t-1}}+ \frac{\beta_3}{1-\beta_3} \frac{S_{t-1}^2}{\sqrt{\hat{V}_t}} +
\frac{\alpha}{1-\beta_3} \cdot \frac{\hat{S}^2}{\sqrt{\hat{V}_t}} 
\end{gathered}
\end{equation}

Inferring from \textbf{Lemma} 10.4 and \textbf{Theorem} 10.5 in Kingma's work and \textbf{Theorem} 4 in Reddi's work~\cite{reddi2019convergence}, using a sequence $\lbrace 1, 2, \cdots, T\rbrace$ to replace $t$ in \eqref{eqA8} to generate $t+1$ equations, and calculate the summation of these equations, we have:
\begin{equation} \tag{A9} \label{eqA9}
\begin{gathered}
\Sigma_{t=1}^T g_t^3(x_t - x_T) \leq \Sigma_{i=1}^D \frac{1}{2\alpha (1-\beta_3)} (x_1-x_T)^2 \sqrt{\hat{V}_{1,i}} +\\
\frac{1}{2(1-\beta_3)}\Sigma_{i=1}^D \Sigma_{t=2}^T (\frac{\sqrt{\hat{V}_{t,i}}}{\alpha}-\frac{\sqrt{\hat{V}_{t-1,i}}}{\alpha}) + \Sigma_{i=1}^D \Sigma_{t=1}^T (x_t-x_t)^2 \sqrt{\hat{V}_{t,i}}\\+ K_3 \Sigma_{i=1}^D \left \| g_{1:t,i}\right \|^2 \\
K_3<\infty
\end{gathered}
\end{equation}

Inferring from \textbf{Theorem} 10.5 in Kigma's work~\cite{Kingma2014adam} and \textbf{Theorem} 4 in Reddi's work~\cite{reddi2019convergence}, we have:
\begin{equation} \tag{A10} \label{eqA10}
\begin{gathered}
\Sigma_{t=1}^T g_t^3(x_t - x_T) \leq \frac{K_1^2}{2\alpha(1-\beta_3)}\Sigma_{i=1}^D \sqrt{T\hat{V}_{T,i}} + \frac{K_2}{2\alpha} \Sigma_{i=1}^D \Sigma_{t=1}^T \frac{\beta_{3,t}}{(1-\beta_{3,t})} \sqrt{t\hat{V}_t}+\\
K_3 \Sigma_{i=1}^D \left \| g_{1:t,i}\right \|^2\\
K_1, K_2, K_3 < \infty
\end{gathered}
\end{equation}

Furthermore, we use a sequence $\lbrace 1, 2, \cdots, T-1\rbrace$ to replace $t$ in Eq.~\eqref{eqA10} and calculate the sum of these equations. According to \textbf{Assumption}~\ref{assumption1}, we can infer:
\begin{equation} \tag{A11} \label{eqA11}
\begin{gathered}
\Sigma_{t=1}^{T-1}(f(x_t)-f(x_T)) \leq \Sigma_{t=1}^{T-1} g_t \cdot (x_t-x_{t+1})
\end{gathered}
\end{equation}

According to \textbf{Assumption}~\ref{assumption4} and \textit{Intermediate Value Theorem}, we have:
\begin{equation} \tag{A12} \label{eqA12}
\begin{gathered}
\Sigma_{t=1}^{T-1} g_t^3 \cdot (x_t-x_{t+1}) = g^3\\
\end{gathered}
\end{equation}

Inferring from Eqs.~\eqref{eqA10} and~\eqref{eqA12}, we conclude: 
\begin{equation} \tag{A13} \label{eqA13}
\begin{gathered}
\left \| g \right \| \leq (\left \| \frac{K_1^2}{2\alpha (1-\beta_3)}\Sigma_{i=1}^D \sqrt{T\hat{V}_{T,i}} + \frac{K_2^2}{2\alpha} \Sigma_{i=1}^D \Sigma_{t=1}^T \frac{\beta_{3,t}}{(1-\beta_{3,t})} \sqrt{t\hat{V}_t} + K_3 \Sigma_{i=1}^D \left \| g_{1:t,i} \right \|^2 \right \|)^{\frac{1}{3}} \\
K_1, K_2, K_3 < \infty
\end{gathered}
\end{equation}

Inferring from Eq.~\eqref{eqA10}, considering $T$ is sufficiently large, we have:
\begin{equation} \tag{A14} \label{eqA14}
\begin{gathered}
\left \| g \right \| = O ( T^{1 \slash 6} )
\end{gathered}
\end{equation}

Let $\left \| \Sigma_{t=1}^{T-1}(f(x_t)-f(x_T)) \right \|$ be $RES$. Inferring from Eq.~\eqref{eqA14} and \textbf{Assumption} \ref{assumption4}, we have:
\begin{equation} \tag{A15} \label{eqA15}
\begin{gathered}
\frac{RES}{T} \leq  \frac{\left \| \Sigma_{t=1}^{T-1} g_t \cdot (x_t-x_T) \right \|}{T} = \frac{\left \|\eta g \right \|}{T} = O(1 \slash T^{5 \slash 6})
\end{gathered}
\end{equation}

Finally, we conclude:
\begin{equation} \tag{A16} \label{eqA16}
\begin{gathered}
\frac{\left \| RES \right \|}{T} = O(\frac{1}{T^{\frac{5}{6}}})
\end{gathered}
\end{equation}
It demonstrates that the convergence bound of \textit{HOME}-3 is $O(\frac{1}{T^{\frac{5}{6}}})$ when $\left \| g_t - g \right \| < \epsilon, \forall \epsilon>0$ and $\left \| g_t \right \|$ is sufficiently large. The following proof demonstrates that the convergence bound could be reduced when the gradient norm $\left \| g_t \right \|$ becomes smaller at the late stage.
\\[0.25cm]
2). On the other hand, we investigate the convergence bound when $\left \| g_t \right \|< 1$ for any $t$.

We can infer from \textbf{Assumption}~\ref{assumption1} and \eqref{eqA16}. Then we have:
\begin{equation} \tag{A17} \label{eqA17}
\begin{gathered}
\frac{RES}{T} \leq \frac{K_1^2}{2\alpha (1-\beta_3)}\Sigma_{i=1}^D \sqrt{T\hat{V}_{T,i}} + \frac{K_2^2}{2\alpha} \Sigma_{i=1}^D \Sigma_{t=1}^T \frac{\beta_{3,t}}{(1-\beta_{3,t})} \sqrt{t\hat{V}_t}+\\
K_3 \Sigma_{i=1}^D \left \| g_{1:t,i}\right \|^2\\
K_1, K_2, K_3 < \infty \\
\end{gathered}
\end{equation}

Similarly, suppose $T$ is sufficiently large, we can conclude:
\begin{equation} \tag{A18} \label{eqA18}
\begin{gathered}
\frac{\left \| RES \right \|}{T} = O(\frac{1}{T^{\frac{1}{2}}})
\end{gathered}
\end{equation}

We have proved \textbf{Theorem}~\ref{theorem1}. \textbf{Theorem}~\ref{theorem1} demonstrate that \textit{HOME}-3 can provide the convergence upper bound between $O(\frac{1}{T^{\frac{1}{2}}})$ and $O(\frac{1}{T^{\frac{5}{6}}})$. To summarize, the beginning gradient is usually large, \textit{HOME}-3 provides a better convergence bound approximately to $O(\frac{1}{T^{\frac{5}{6}}})$. In the late stage, with the norm of gradient gradually reduced, the convergence bound of \textit{HOME}-3 decreases to $O(\frac{1}{T^{\frac{1}{2}}})$. The performance of \textit{HOME}-3 is comparable to Adam~\cite{Kingma2014adam} in the late stage, such as the gradient getting stuck in a stationary point.

\textbf{Theorem 4.2} Let $f$ satisfy \textbf{\textit{Assumption}} 2, suppose $T$ as the maximum iteration, inferring from \textbf{\textit{Definitions}} 3, 5, and 6, then $\frac{\left \| f(x_0)-f(x_T) \right \|}{T} = O(1 \slash T^{5 \slash 6})$ holds.

\textit{\textbf{Proof}}:\\
1) At the early stage, the norm of gradient $g_t$ is sufficiently large, and the higher-order momentum using $g_t^3$ dominates the update, such as $\vert \vert g_t^3 \vert \vert >> \vert \vert g_t \vert \vert$.

According to \textbf{Assumption}~\ref{assumption2}, we have:
\begin{equation} \tag{A19} \label{eqA19}
\begin{gathered}
f(x_{t+1})-f(x_t) \leq g_t (x_{t+1}-x_t)+\frac{L}{2}(x_{t+1}-x_t)(x_{t+1}-x_t)^T
\end{gathered}
\end{equation}

Since $(x_{t+1}-x_t)$ and $(x_{t+1}-x_t)^T$ are bounded, we let 
\begin{equation} \tag{A20} \label{eqA20}
\begin{gathered}
\left \| (x_{t+1}-x_t)(x_{t+1}-x_t)^T \right \| \leq K_M \left \| (x_{t+1}-x_t) \right \|
\end{gathered}
\end{equation}

Next, we use a sequence $\lbrace 1, 2, \cdots, T-1 \rbrace$ to replace $t$ in Eq.~\eqref{eqA16} and calculate the sum of these equations. We can infer:
\begin{equation} \tag{A21} \label{eqA21}
\begin{gathered}
\left \| f(x_1)-f(x_T) \right \| \leq \left \| \Sigma_{t=1}^{T-1} g_t \cdot (x_{t+1}-x_t) + \frac{L}{2} \cdot (x_T-x_1)\right \|
\end{gathered}
\end{equation}

According to \textbf{Definition} \ref{def2}, $L<\infty$, thus, $\left \| f(x_1)-f(x_T) \right \|$ only relates to term $\left \|\Sigma_{t=1}^{T-1} g_t \cdot (x_{t+1}-x_t) \right \|$.

Since $\left \| g_t^3 \right \| >> \left \| g_t \right \|$, $ \forall t \in \lbrace 1,t\rbrace$, we can infer:
\begin{equation} \tag{A22} \label{eqA22}
\begin{gathered}
\left \|\Sigma_{t=1}^{T-1} g_t \cdot (x_{t+1}-x_t) \right \| \leq  \left \| g_t^3 \right \| \cdot \left \|\Sigma_{t=1}^{T-1}  (x_{t+1}-x_t) \right \|
\end{gathered}
\end{equation}

According to Eqs.~\eqref{eqA20},~\eqref{eqA21}, and~\eqref{eqA22} as well as \textbf{Theorem}~\ref{theorem1}, under \textbf{Assumption} \ref{assumption2}, similarly, we can conclude:
\begin{equation} \tag{A23} \label{eqA23}
\begin{gathered}
\frac{\left \| f(x_1)-f(x_T) \right \|}{T} \leq  \frac{1}{T} \cdot \left \| \Sigma_{t=1}^{T-1} g_t \cdot (x_t-x_{t+1}) \right \| + \frac{K_M}{2T}
\end{gathered}
\end{equation}

Since we previously proved $\left \| g_t \right \| = O(T^{\frac{1}{6}})$, suppose $T$ is sufficiently large, we can infer $\frac{1}{T} \cdot \left \| \Sigma_{t=1}^{T-1} g_t \cdot (x_t-x_{t+1}) \right \|$ is equal to $O(\frac{1}{T^{\frac{5}{6}}})$.

Thus, the convergence bound of \textit{HOME}-3 is $O\left(\frac{1}{T^{5/6}}\right)$, assuming the norm of gradient is sufficiently large at initialization.

On the other hand, considering the norm of gradient is not sufficiently large. In that case, the lower-order momentum using $g_t$ can dominate the process, such as $\vert \vert g_t \vert \vert >> \vert \vert g_t^3 \vert \vert$

Similar to Eqs.~\eqref{eqA22} and~\eqref{eqA23}, we can infer:
\begin{equation} \tag{A24} \label{eqA24}
\begin{gathered}
\frac{\left \| f(x_1)-f(x_T) \right \|}{T} \leq \left \| \Sigma_{t=1}^{T-1} g_t \cdot (x_t-x_{t+1}) \right \| +\frac{K_M}{T}
\end{gathered}
\end{equation}

Since $\frac{1}{T} \cdot \left \| \Sigma_{t=1}^{T-1} g_t \cdot (x_t-x_{t+1}) \right \|=O(\frac{1}{T^{\frac{1}{2}}})$, we proved that \textit{HOME}-3 can obtain convergence bound $O(\frac{1}{T^{\frac{1}{2}}})$ when the norm of gradient is not large.

In conclusion, \textit{HOME}-3 can provide a comparable convergence bound under the smooth nonconvex Assumption (please refer to \textbf{Assumption}~\ref{assumption2}). The only potential issue is the smoothness of the objective function. If $L>>T$ in Eq.~\eqref{eqA21}, the convergence bound could be seriously influenced.

\textbf{Lemma 4.3} (Norm of Coordinate Randomization Operator is Equal to 1) Suppose the permutation randomization as an operator $\mathcal{R}:\mathbb{R}^D \rightarrow \mathbb{R}^D$, $\left \| \mathcal{R} \right \| =1$ holds, if $D < \infty$.

\textbf{\textit{Proof}}:\\
Considering $\mathcal{R}$ applying on finite-dimensional space:
\begin{equation} \tag{A25} \label{eqA25}
\begin{gathered}
\mathcal{R} \cdot \left[ 
\begin{array}{c}
     x_1\\
     x_2\\
     \vdots\\
     x_D\\
     \end{array}
     \right]
 =  \left[ \begin{array}{c}
     \hat{x}_1\\
     \hat{x}_2 \\
     \vdots\\
     \hat{x}_D\\
     \end{array}
     \right]
\end{gathered}
\end{equation}

Inferring from Eq.~\eqref{eqA13}, we have:
\begin{equation} \tag{A26} \label{eqA26}
\begin{gathered}
 \hat{x}_1 = x_i, \hat{x}_2 = x_j, \cdots, \hat{x}_D = x_k, i,j,k \in \lbrack 1,D \rbrack
\end{gathered}
\end{equation}

Inferring from Eq.~\eqref{eqA26}, we have:
\begin{equation} \tag{A27} \label{eqA27}
\begin{gathered}
 \vert \vert \lbrace x_1, x_2, \cdots, x_D \rbrace \vert \vert = \vert \vert \lbrace \hat{x}_1, \hat{x}_2, \cdots, \hat{x}_D \rbrace \vert \vert
\end{gathered}
\end{equation}

According to the concept of operator norm \citep{Rudin1973functionalanalysis}, we can derive the following:
\begin{equation} \tag{A28} \label{eqA28}
\begin{gathered}
 \vert \vert \mathcal{R} \vert \vert = sup \frac{\mathcal{R} \cdot \vert \vert \lbrace x_1, x_2, \cdots, x_D \rbrace \vert \vert}{\vert \vert \lbrace x_1, x_2, \cdots, x_D \rbrace \vert \vert} = sup \frac{\vert \vert \lbrace \hat{x}_1, \hat{x}_2, \cdots, \hat{x}_D \rbrace \vert \vert}{\vert \vert \lbrace x_1, x_2, \cdots, x_D \rbrace \vert \vert}=1
\end{gathered}
\end{equation}

\textbf{Theorem 4.4} (Coordinate Randomization Maintains The Convergence Bound of Incorporated Optimizer) Inferring from \textbf{Lemma} \ref{lemma1}, the convergence bound of a gradient-based optimizer incorporating coordinate randomization $\mathcal{R} \cdot \mathcal{G}$ should be equal to the convergence bound of an original gradient-based optimizer $\mathcal{G}$ without coordinate randomization.

\textbf{\textit{Proof}}:\\
Inferring from the concept of contraction operator~\citep{Rudin1973functionalanalysis}, we have:
\begin{equation} \tag{A29} \label{eqA29}
\begin{gathered}
 \vert \vert \mathcal{G} \cdot (f(X)-f(Y)) \vert \vert \leq c \vert \vert \mathcal{G} \cdot (f(X)-f(Y)) \vert \vert\\
 0<c<1
\end{gathered}
\end{equation}

We can rewrite the left side of Eq.~\eqref{eqA16} as:
\begin{equation} \tag{A30} \label{eqA30}
\begin{gathered}
 \vert \vert \mathcal{G} \cdot (f(I_{t+1})-f(I_t)) \vert \vert
\end{gathered}
\end{equation}

Then, we have:
\begin{equation} \tag{A31} \label{eqA31}
\begin{gathered}
 \vert \vert \mathcal{G} \cdot (f(I_{t+1})-f(I_t)) \vert \vert \leq c \cdot \vert \vert (f(I_{t+1})-f(I_t)) \vert \vert
\end{gathered}
\end{equation}

Considering the incorporation of optimizer and randomization as$\mathcal{R} \cdot \mathcal{G} \cdot f(x)$, we have
\begin{equation} \tag{A32} \label{eqA32}
\begin{gathered}
 \vert \vert \mathcal{R} \cdot \mathcal{G} \cdot (f(I_{t+1})-f(I_t)) \vert \vert \leq \vert \vert \mathcal{R} \vert \vert \cdot \vert \vert \mathcal{G} \cdot(f(I_{t+1})-f(I_t)) \vert \vert
\end{gathered}
\end{equation}

Inferring from \textbf{Lemma} \ref{lemma1}, it is obvious that we have:
\begin{equation} \tag{A33} \label{eqA33}
\begin{gathered}
\vert \vert  \mathcal{R} \vert \vert \cdot \vert \vert \cdot \mathcal{G} \cdot (f(I_{t+1})-f(I_t)) \vert \vert =  \vert \vert \mathcal{G} \cdot(f(I_{t+1})-f(I_t)) \vert \vert \leq c \cdot \vert \vert f(I_{t+1}) -f(I_t) \vert \vert
\end{gathered}
\end{equation}

Eq.~\eqref{eqA33} implies permutation randomization $\mathcal{R}$ can maintain the convergence rate of original gradient-based optimizer $\mathcal{G}$.

\textbf{Additional Experimental Results}:

In additional experiments, we compare the time consumption of \textit{HOME}-3 with other peer optimizers.
\begin{table}[H]
  \caption{Time Consumption Comparison in Seconds of \textit{HOME}-3 and Other Peer Three Optimizers}
  \centering
  \begin{tabular}{cc}
    \toprule               \\
   Time Consumption at 1st Layer     &  Time Consumption at 2nd Layer  \\
    \midrule
    ADMM $431.58 \pm 83.56$     & ADMM $247.42 \pm 68.54$       \\
    ADAM  $961.65 \pm 199.67$   & ADAM  $585.37 \pm 55.17$      \\
    STORM  $4711.35 \pm 342.25$  & STORM   $4616.66 \pm 556.27$         \\
    HOME-3 $1262.66 \pm 195.16$  & HOME-3 $1108.62 \pm 188.05$ \\
    \bottomrule
  \end{tabular}
  \label{table:table2}
\end{table}

\begin{figure}H]
  \centering
  \includegraphics[width=0.96\textwidth]{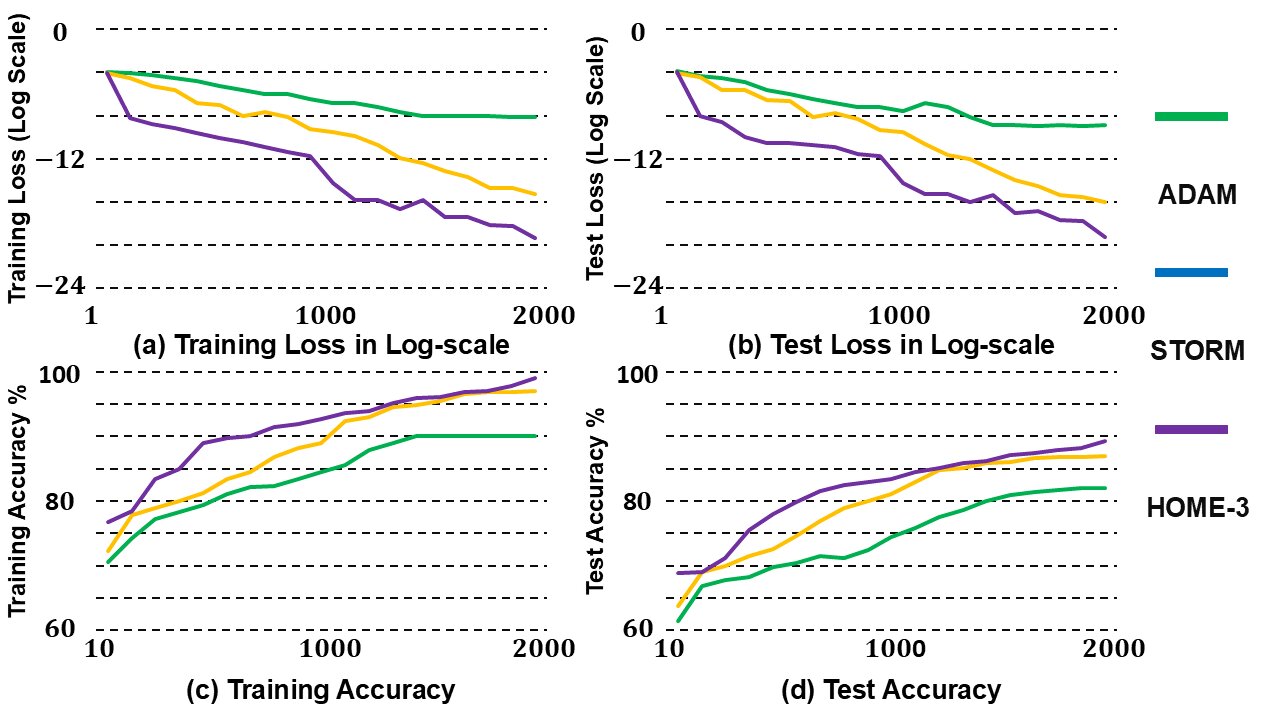}  \caption{An illustration of reconstruction loss comparisons of \textit{HOME}-3 and other peer optimizers on solving logistic regression problem.}
  \label{fig:fig5}
\end{figure}

Furthermore, to ensure a fair comparison among different methods for optimizing supervised learning problems, we set all parameters to reported default values~\cite{Kingma2014adam, cutkosky2019momentum}. Each method was then employed to solve a logistic regression problem \citep{schober2021logistic} using publicly released breast cancer data~\cite{BreastCancer} for classification. The results, observed within iterations 1 to 2000, are illustrated in Figure~\ref{fig:fig5}.

Moreover,  we present a comparison of reconstruction errors using a three-layer DBN, illustrating a representative case of nonsmooth and nonconvex optimization. The maximu iteration is 6000.
\begin{figure}[H]
  \centering
  \includegraphics[width=0.96\textwidth]{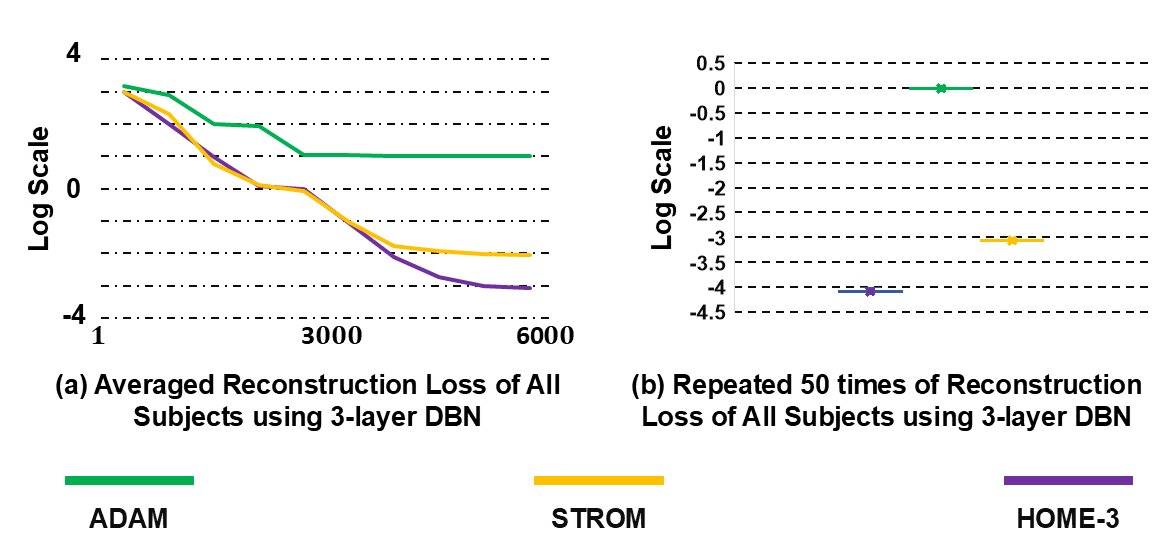}
  \caption{An illustration of reconstruction loss comparisons of \textit{HOME}-3 and other peer optimizers on optimizing 3-layer DBN.}
  \label{fig:fig6}
\end{figure}

\end{document}